\documentclass[preprint,12pt]{elsarticle}

\usepackage[utf8]{inputenc}
\usepackage{amssymb}
\usepackage{url}
\usepackage{amsmath}
\usepackage{graphicx}%
\usepackage{multirow}%
\usepackage{amsmath,amssymb,amsfonts}%
\usepackage{amsthm}%
\usepackage{mathrsfs}%
\usepackage[title]{appendix}%
\usepackage{xcolor}%
\usepackage{textcomp}%
\usepackage{manyfoot}%
\usepackage{booktabs}%
\usepackage{algorithm}%
\usepackage{algorithmicx}%
\usepackage{algpseudocode}%
\usepackage{listings}%
\usepackage{comment}
\usepackage{tabularx} 
\usepackage{array}
\usepackage{pdflscape}
\usepackage{booktabs}
\usepackage{siunitx}
 \usepackage{amsmath}
 \usepackage{lineno}

\journal{Medical Image Analysis}

\begin{document}

\begin{frontmatter}

\title{A Speech-to-Video Synthesis Approach Using Spatio-Temporal Diffusion for Vocal Tract MRI}

\author[1,2,3]{Paula Andrea Pérez-Toro\corref{cor1}}
\ead{paula.andrea.perez@fau.de}
\cortext[cor1]{Corresponding Author}

\author[1,2,3]{Tomás Arias-Vergara}

\author[1]{Fangxu Xing}

\author[4]{Xiaofeng Liu}

\author[5,6]{Maureen Stone}

\author[5]{Jiachen Zhuo}

\author[2,3]{Juan Rafael Orozco-Arroyave}

\author[2]{Elmar Nöth}

\author[7]{Jana Hutter}

\author[8]{Jerry L. Prince}

\author[2]{Andreas Maier}

\author[1]{Jonghye Woo}

\affiliation[1]{organization={Harvard Medical School/Massachusetts General Hospital}, city={Boston}, postcode={02114}, state={Massachusetts}, country={USA}}

\affiliation[2]{organization={Pattern Recognition Lab, Department of Computer Science, Friedrich-Alexander-Universität Erlangen-Nürnberg}, city={Erlangen}, postcode={91058}, state={Bayern}, country={Germany}}

\affiliation[3]{organization={GITA Lab, Faculty of Engineering, Universidad de Antioquia}, city={Medellín}, postcode={050010}, state={Antioquia}, country={Colombia}}

\affiliation[4]{organization={Department of Radiology \& Biomedical Imaging and Biomedical Informatics \& Data Science, Yale University}, city={New Heaven}, postcode={06510}, state={Connecticut}, country={USA}}

\affiliation[5]{organization={Department of Neural and Pain Sciences and Department of Orthodontics and Pediatrics, University of Maryland School of Dentistry}, city={Baltimore}, postcode={21210}, state={Maryland}, country={USA}}

\affiliation[6]{organization={Department of Orthodontics and Pediatrics, University of Maryland School of Dentistry}, city={Baltimore}, postcode={21201}, state={Maryland}, country={USA}}

\affiliation[7]{organization={Smart Imaging Lab, Radiological Institute, University Hospital Erlangen}, city={Erlangen}, postcode={91052}, state={Bayern}, country={Germany}}

\affiliation[8]{organization={Department of Electrical and Computer Engineering, Johns Hopkins University}, city={Baltimore}, postcode={21218}, state={Maryland}, country={USA}}

\begin{abstract}
Understanding the relationship between vocal tract motion during speech and the resulting acoustic signal is crucial for aided clinical assessment and developing personalized treatment and rehabilitation strategies. Toward this goal, we introduce an audio-to-video generation framework for creating  Real Time/cine-Magnetic Resonance Imaging (RT-/cine-MRI) visuals of the vocal tract from speech signals. Our framework first preprocesses RT-/cine-MRI sequences and speech samples to achieve temporal alignment, ensuring synchronization between visual and audio data. We then employ a modified stable diffusion model, integrating structural and temporal blocks, to effectively capture movement characteristics and temporal dynamics in the synchronized data. This process enables the generation of MRI sequences from new speech inputs, improving the conversion of audio into visual data. We evaluated our framework on healthy controls and tongue cancer patients by analyzing and comparing the vocal tract movements in synthesized videos. Our framework demonstrated adaptability to new speech inputs and effective generalization. In addition, positive human evaluations confirmed its effectiveness, with realistic and accurate visualizations, suggesting its potential for outpatient therapy and personalized simulation of vocal tract visualizations.
\end{abstract}



\end{frontmatter}



\section{Introduction}

Understanding the complex dynamics of the vocal tract using Magnetic Resonance Imaging (MRI) has the potential for diagnosing and effectively treating speech-related disorders~\cite{ziegler2019motor,richmond2012ultrax,scott2014speech}. From both patient and therapist perspectives, however, the use of MRI to monitor and treat speech and vocal tract disorders comes with notable limitations~\cite{mclean2023mri}. For example, patients often find it challenging to make frequent visits to the clinic for MRI scans due to logistical issues, such as travel difficulties, cost, time constraints, and the physical and psychological stress of undergoing the procedure. For therapists, these limitations restrict the ability to conduct continuous, real-time assessment of treatment progress, potentially delaying adjustments to therapy that could enhance patient outcomes. To bridge this gap, we introduce a framework designed to convert speech into MRI sequences. Incorporating speech-to-video synthesis of the vocal tract visualization into medical practice not only bridges practical gaps but also potentially represents a step forward in the analysis and simulation of speech and vocal tract dynamics.

Recent advances in text-to-video synthesis have been driven by the development of Diffusion Probabilistic Models (DPMs)~\cite{ho2020denoising}, outperforming traditional models, such as Generative Adversarial Networks (GANs)~\cite{goodfellow2020generative} and Variational Autoencoders (VAEs)~\cite{kingma2019introduction} in creating images with greater diversity and fidelity~\cite{dhariwal2021diffusion}. Developments in sampling techniques and the employment of spaces with lower dimensionality through Latent Diffusion Models (LDM)~\cite{yang2023diffusion}, have addressed efficiency and scalability challenges. The integration with Transformer models in natural language processing, such as CLIP~\cite{radford2021learning}, has improved the alignment between text and generated images, leading to the production of high-fidelity videos. Techniques ranging from video diffusion to leveraging pre-trained models such as DALL-E~\cite{luo2023videofusion} and SORA~\cite{sora} have further refined the synthesis process, enabling the creation of realistic videos directly from textual prompts, thereby opening new possibilities for content generation and multimedia applications.


In addition, speech-driven talking face synthesis has evolved through speaker-specific~\cite{wang2022one,mildenhall2021nerf} and speaker-independent models~\cite{wu2023speech2lip}. While speaker-specific methods have produced realistic visuals with less video data required~\cite{guo2021ad}, challenges persist in accurately predicting aspects such as morphological properties, which are crucial in clinical scenarios. These methods leverage Neural Radiance Fields (NeRF)~\cite{mildenhall2021nerf} to enhance learning from short videos. However, issues with motion synchronization remain, highlighting the ongoing need for innovation in achieving realistic and synchronized talking face synthesis. In~\cite{liu2024tagged} the authors proposed a framework for synthesizing cine-MRI from tagged MRI sequences, where the use of light spatial-temporal transformer and a recurrent sliding fine-tuning scheme, the method enhances both spatial and temporal consistency. Complementing these advancements, in~\cite{liu2024tagged}, the authors introduce a framework for synthesizing cine-MRI from tagged MRI sequences of vocal track visualizations that employs a light spatial-temporal transformer and a recurrent sliding fine-tuning scheme to enhance spatial and temporal consistency, demonstrating its potential in improving medical imaging techniques.

In the context of speech synthesis from MRI sequences, recent research has employed various techniques, such as encoder-decoder networks for predicting vocal tract articulator contours~\cite{ribeiro2022automatic}, CNN-BiLSTM models for mel-spectrogram prediction~\cite{otani2023speech}, and a plastic light Transformer framework for translating tongue movement weighting maps from non-negative matrix factorization into audio waveforms~\cite{liu2023speech}. In addition, most of the approaches in the literature based on diffusion models aim to translate spoken language into corresponding videos/images. However, some studies have also addressed the challenge of generating coherent video translations from speech. This domain focuses on converting spoken language into corresponding video content, including the realistic animation of human faces, body movements, and even the generation of entire scenes that align with the spoken narrative. The study in~\cite{nguyen2024speech2rtmri} also explored a similar area, using speech-to-video models to dynamically generate Real Time (RT) MRI videos of the vocal tract reacting to spoken inputs. While their approach marks an important step forward, it tends to produce videos that could be improved in terms of articulatory detail and transition smoothness.

In this work, we leverage generative models based on LDMs~\cite{yang2023diffusion} to create realistic and dynamic visual simulations of the vocal tract's response to speech sounds, focusing on textless generation. This approach provides not only a deeper understanding of individual speech pathologies but also paves the way for personalized treatment strategies. By offering an engaging and accessible therapeutic process, it holds significant potential to enhance patient outcomes and accessibility in speech therapy.

Our method builds upon spatio--temporal stable diffusion models, advancing beyond prior techniques such as Vector Quantized GAN (VQGAN), Diffusion AutoEncoder (DiffAE), and traditional stable diffusion models, which predominantly focus on spatial information. By operating within a diffusion probabilistic framework in the latent space, our approach captures intricate spatio-temporal dynamics, ensuring superior temporal coherence and fine spatial detail preservation. These models are scalable for high-dimensional video data and integrate audio embeddings as conditioning inputs, enabling precise alignment between speech signals and generated visualizations. This makes our approach particularly effective for synthesizing RT- and cine-MRI visualizations of the vocal tract, offering a robust solution for personalized and dynamic speech-related simulations.

\section{Methods}
The methodologies addressed in our study focused on generative-based models that are able to synthesize and manipulate images. 
\sloppy
We introduce Speech2-MRI, a spatio--temporal latent diffusion-based model designed for speech-to-video translation. To determine the efficacy of image synthesis methods in reproducing the corresponding speech sequences, we also used the VQGAN and DiffAE with conditioning models for comparison.

\subsection{Data}

For evaluation purposes, we used two distinct datasets that provide RT- or cine-MRI visualizations of the vocal tract. All participants provided their informed consent to participate.

\subsubsection{USC 75-Speaker Speech Real Time MRI Database}
The first database is the publicly available \textit{USC 75-Speaker Speech MRI Database}~\cite{lim2021multispeaker}, which is a collection of 2D sagittal-view RT-MRI videos, paired with synchronized audio recordings. This dataset includes contributions from 75 participants, comprising both native and non-native American English speakers, balanced across genders, and ranging in age from 18 to 59 years. Each participant performed 21 distinct speech tasks, resulting in 32 recordings per subject. The speech tasks were designed to cover a broad spectrum of phonetic and phonological phenomena, including the production of vowels and consonants, reading passages, and spontaneous speech tasks such as describing pictures. Overall, the database encompasses roughly 21 hours of recording, amounting to approximately 6,350,000 frames in total. The data were collected using a commercial 1.5 Tesla MRI scanner, outfitted with a custom 8-channel upper airway receiver coil array. This coil includes four elements on each side of the subject's cheeks, specifically designed to enhance signal reception and improve the signal-to-noise ratios for capturing speech-related structures.

\subsubsection{Tongue Cancer cine-MRI}
For the second in-house dataset, we acquired paired 2D cine-MRI sequences and synchronized audio recordings from 30 participants, comprising 10 individuals diagnosed with tongue cancer and 20 healthy controls. The subjects were instructed to articulate the phrases ``a souk'' and ``a geese'' rhythmically, guided by a metronome-like auditory cue to maintain consistent timing. Each MRI sequence contained 26 frames, resized to a uniform resolution of $128\times128$ for uniformity. The data acquisition was conducted using a Siemens 3.0T TIM Trio MRI system, equipped with a 12-channel head coil and a 4-channel neck coil. A segmented gradient echo sequence was employed, yielding a field of view of $240\times240$ mm with an in-plane resolution of $1.87\times1.87$ mm and a slice thickness of 6 mm. The temporal resolution was 36 milliseconds, ensuring no delays during the one-second acquisition period for each task. Collecting synchronized cine-MRI, and audio data presented significant challenges due to the prohibitive nature of metal devices in MRI environments, which necessitated meticulous planning and specialized equipment. However, this synchronized multimodal dataset provides valuable insights for analyzing speech and tongue motion dynamics in both healthy individuals and patients with tongue cancer.

\subsection{Preprocessing and Training}
\begin{figure}[!htpb]
    \centering
    \includegraphics[width=\linewidth]{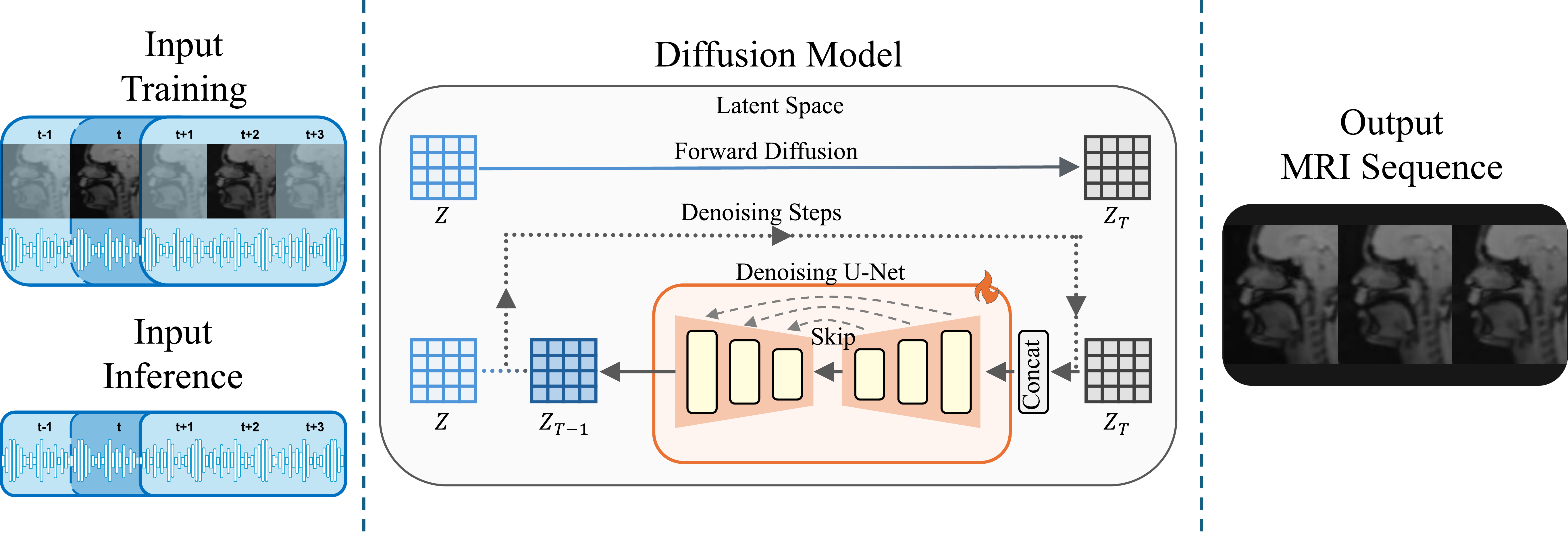}
    \caption{Methodology proposed in this study, where a diffusion model designed for generating MRI sequences, detailing its workflow in three main stages. Initially, MRI frames and corresponding audio waveforms are input for training, while for inference only the audio is considered. In the diffusion phase, these inputs are transformed into a latent space and progressively noised before being incrementally denoised. The process culminates in the output of an MRI sequence.}
    \label{fig:metho1}
\end{figure}

The general methodology for the spatial and spatio-temporal models is displayed in Fig.~\ref{fig:metho1}.

To synchronize the speech samples with the images, we align them temporally, using a video frame rate of $26\,fps$ and processing the speech at a sampling frequency of 16 kHz. This is done so that images with a resolution of 128$\times$128 pixels correspond to approximately $39\,ms$ of speech for each image frame.

\begin{figure}[!htpb]
    \centering
    \includegraphics[width=\linewidth]{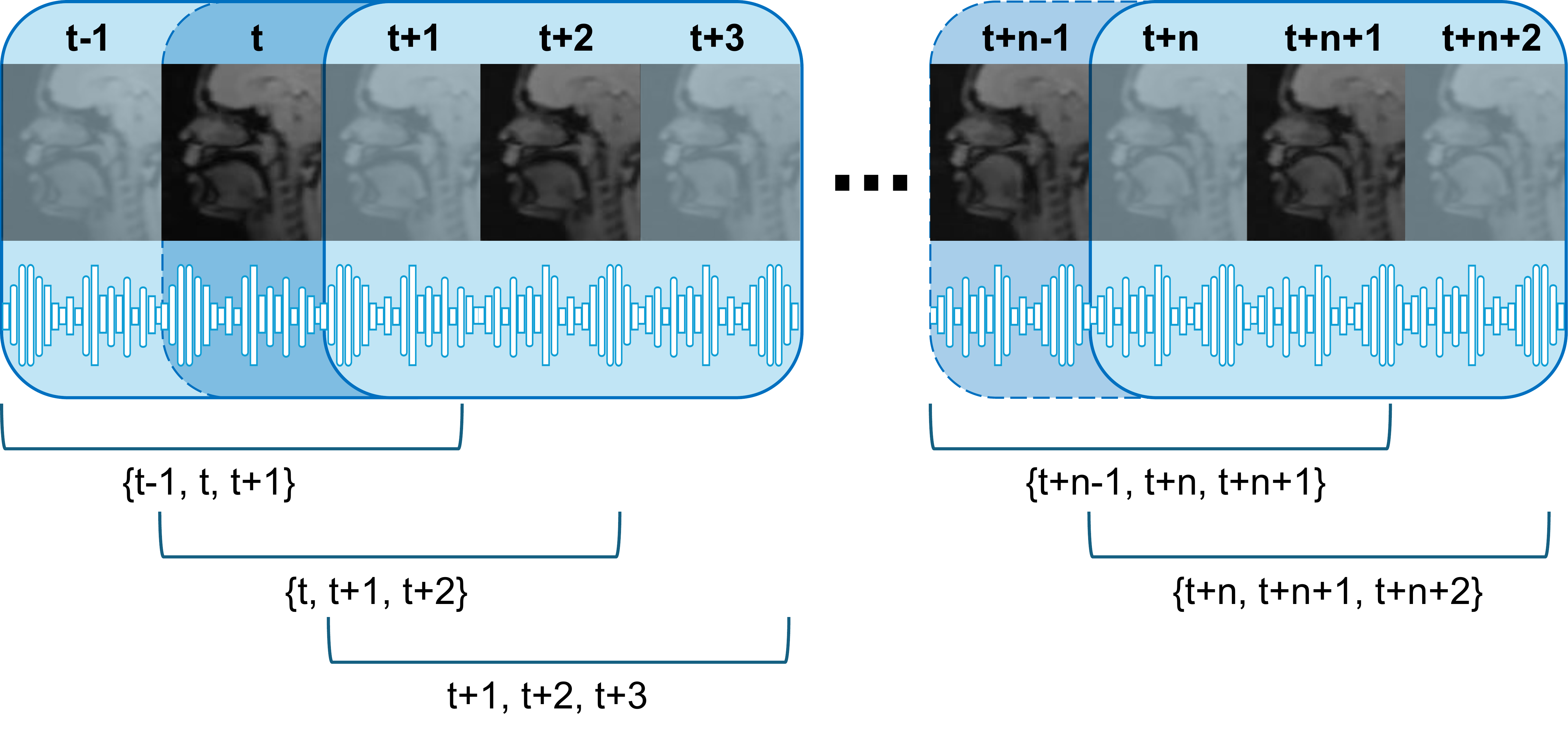}
    \caption{Temporal alignment of speech samples and image frames, where it is illustrated how overlapping windows of images and speech segments are synchronized for effective analysis. Each image frame corresponds to approximately 67\,ms of speech. To capture temporal context, bidirectional segments spanning three consecutive windows (approximately 200\,ms) are used.}
    \label{fig:wind_seg}
\end{figure}

To capture temporal context from both preceding and subsequent intervals, we use bidirectional segments covering three windows, which corresponds to approximately $115\,ms$ (see Fig.~\ref{fig:wind_seg}). This approach not only incorporates temporal dynamics but also mitigates the impact of audio-visual misalignment observed in the datasets used.

In the case of spatial models such as VQGAN and DiffAE, training occurs frame by frame, with the ability to process sequences derived from mapping specific speech segments to each frame. Conversely, the STDiff model employs a different strategy to handle spatio--temporal information. It not only associates speech with individual frames but also ensures smooth transitions by processing three frames at a time, each corresponding to a $200\,ms$ speech segment. This approach allows the model to incorporate both past and future auditory information, as well as the preceding and subsequent visual data of the frames, thus creating a more integrated and fluid representation of movement and sound.

 Additionally, we implemented our framework using PyTorch and trained it on four NVIDIA H1000 GPUs, during 10000 epochs for the pretraining, 3000 for fine-tuning of the \textit{USC 75-Speaker} dataset, and 1000 epochs for the second dataset. For these experiments, the learning rate was set to $5e^{-5}$, the time steps for the schedulers were set to 1000 for training and to 25 in inference.

\subsection{Encoding Audio with Wav2Vec 2.0}

We leverage the feature embeddings extracted by Wav2Vec 2.0 to condition the diffusion processes. This algorithm processes raw audio inputs to produce encoded representations that are widely used in speech recognition systems. The Wav2Vec 2.0 architecture comprises three key components: a feature extraction module, a context network, and a projection layer for output generation. During the feature extraction phase, temporal convolutions transform raw speech into a compact latent representation. Specifically, given a sequence of raw audio samples $\mathbf{x} = (x_1, x_2, ..., x_T)$, the convolutional feature encoder $f$ maps $\mathbf{x}$ to latent features $\mathbf{Z}$, where $\mathbf{Z} = (\mathbf{z}_1, \mathbf{z}_2, ..., \mathbf{z}_T)$ represents the encoded feature representations of the input audio.

The model employs an audio-specific adaptation of the Transformer's masking mechanism for self-supervised learning, where segments of audio are masked and quantized. This masking process encourages the model to learn robust representations by predicting missing segments. Thus, contextual representations are generated by combining a Transformer-based method with contrastive learning. These representations are processed by the Transformer encoder $ g: \mathbf{Z} \mapsto \mathbf{C}$, which outputs context representations $ \mathbf{C} = (\mathbf{c}_1, \mathbf{c}_2, ..., \mathbf{c}_T)$. 

During training, parts of the latent representations are masked, and the model is tasked with predicting their quantized counterparts using the context representations as input. Quantized representations are obtained by discretizing the audio space into a finite set of vectors. The training objective minimizes the contrastive loss between the predicted and actual quantized representations of the masked segments.
This approach utilizes only the encoder component of the Transformer, which has been pretrained on 100~hours of speech from native English speakers. For the conditioning of the diffusion process, We use the output of the last hidden layer to encode the information of the raw signal (sequence length = 31 $\times$ embedding dimension = 1024).

\subsection{VQGAN Based Model}

This model is designed to facilitate cross-modal translation, by converting speech audio signals into MRI images, employing a methodology inspired by the architecture proposed in~\cite{esser2021taming}. The model uses Wav2Vec to encode audio inputs, projecting these representations into a 2D format suited for the generator. The generator is then tasked with reconstructing MRI images from these audio-derived representations. At the same time, the discriminator compares both the synthetically generated images and the authentic MRI images to determine the adversarial losses, ensuring the model's output aligns closely with realistic MRI images. We fixed the codebook size to 32 with a latent dimension of 256, which implies that the model quantizes the audio-derived representations into one of 32 possible vectors, each vector being 256-dimensional.
The code for this model was adapted from the implementation of MONAI's VQGAN\footnote{\url{https://github.com/Project-MONAI/MONAI}}.

\subsection{Spatial Diffusion}

The architecture used in this study to compare with the Spatio--Temporal Diffusion (STDiff) is the DiffAE~\cite{preechakul2022diffusion}. It integrates a diffusion autoencoder with semantic encoder based ResNet for achieving meaningful and decodable image representations. It employs a two-part latent code system: a semantically meaningful subcode and a stochastic subcode capturing low-level details for near-exact reconstruction. The key components include a learnable encoder for high-level semantics and a denoising diffusion implicit model as the decoder for modeling stochastic variations. Our system was adapted for grayscale images, using progressive channel expansion, and selective attention mechanisms. 
 The traditional image-based semantic encoder (ResNet-18) is substituted with a Wav2Vec encoder to handle audio data, allowing the system to convert speech-derived features back into their original image formats. 
 This is quantitatively measured using the reconstruction loss ($\mathcal{L}_{\text{reconst}}$), as detailed in Eq.~\ref{eq:loss-rec_diffae}. Each piece of preprocessed speech segments, denoted as $x_{\text{nt}}^i$, is processed by the Wav2Vec encoder $E_{Sp}$, which translates the speech input into a feature representation suitable for the DDIM image decoder $D$. The decoder then attempts to reconstruct the target output, $f_{\text{gt}}^i$, which corresponds directly to the ground truth image.
 In this adapted setup, to ensure seamless integration between the visual and auditory processing streams, the output of the Wav2Vec speech encoder is resized to match the 512-unit output size of the ResNet image encoder. This resizing is accomplished using a linear transformation layer, which modifies the dimensionality of the Wav2Vec output.

\begin{equation}
    \mathcal{L}_{reconst} = \sum_{i=1}^{F} \| D(E_{Sp}(x_{\text{nt}}^i)) - f_{\text{gt}}^i \|^2 
    \label{eq:loss-rec_diffae}
\end{equation}
\newline

Furthermore, contrastive training ($\mathcal{L}_{\text{cont}}$) is used to fine-tune the embedding space alignment between the image ($E_{Im}$) and speech representations ($E_{Sp}$) as defined in Eq.~\ref{eq:loss-cont_diffae}. In this equation, a margin $m$ defines a threshold value beyond which the model is penalized if similar points are too close.  $E_{Im}$ represent the ResNet image encoder. Notice that our primary goal is to bring the embeddings as close together as possible. Therefore, we will assume that all pairs are positive examples, i.e., $y_i=1$.

\begin{equation}
    \begin{split}
        \mathcal{L}_{\text{cont}} &= \sum_{i=1}^{F} \frac{1}{2} \Biggl( y_i \| E_{Im}( f_{\text{gt}}^i) - E_{Sp}(x_{\text{nt}}^i) \|^2 + \\
        &\quad (1 - y_i) \max\left(0, \text{m} - \| E_{Im}(f_{\text{gt}}^i) - E_{Sp}(x_{\text{nt}}^i) \|\right)^2 \Biggr) \\
        &= \sum_{i=1}^{F} \frac{1}{2} \| E_{Im}(f_{\text{gt}}^i) - E_{Sp}(x_{\text{nt}}^i) \|^2
    \end{split}
    \label{eq:loss-cont_diffae}
\end{equation}
\newline

Lastly, the overall loss ($\mathcal{L}_{ov}$) combines these two loss components into a singular training objective (see Eq.~\ref{eq:loss_diffae}).
Here, $\lambda_{1}$ and $\lambda_{2}$ are weighting coefficients that balance the importance of the reconstruction accuracy against the effectiveness of the contrastive learning process.

\begin{equation}
    \mathcal{L}_{ov} =  \lambda_{1} \mathcal{L}_{reconst} +  \lambda_{2} \mathcal{L}_{cont}
    \label{eq:loss_diffae}
\end{equation}
\newline

The code for this model was adapted from the implementation of MONAI's diffusion autoencoder\footnote{\url{https://github.com/Project-MONAI/MONAI}}.

\subsection{Spatio--Temporal Diffusion}

Our proposed architecture for Speech-to-Video translation follows a stable diffusion training pipeline~\cite{rombach2022high} to synthesize coherent video sequences from speech sounds. It integrates spatial and temporal dynamics to ensure seamless frame transitions aligned with the given speech frame. 
The audio embeddings serve as conditioning information for the diffusion process, which iteratively refine the noise into structured video frames $\mathbf{V}$ in $\mathbb{R}^{F \times H \times W \times C}$, where $F$ is the number of frames, $H/W$ are the height/width dimensions, respectively, and $C$ corresponds to the number of channels. This process is facilitated by a conditional 3D-UNet ($G$) that processes both spatial and temporal information~\cite{wang2023modelscope}.
It takes noisy latent representations $\mathbf{f}_{\text{nt}}$ and audio embeddings as inputs at each diffusion timestep $t$ to produce the next latent state $\mathbf{f}_{nt-1}$.

\begin{figure}[!htpb]
    \centering
    \includegraphics[width = \linewidth]{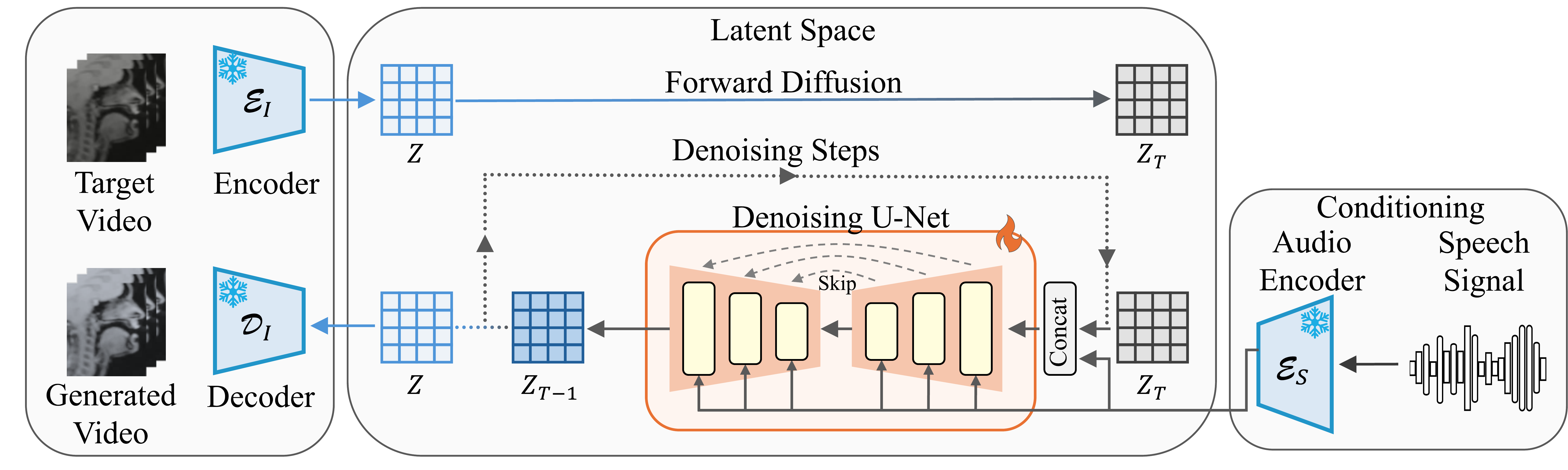}
    \caption{General structure of the spatio--temporal latent diffusion  architecture used to synthesize speech to MRI sequences. It consists of three main steps. (1) Video encoding by using a VAE architecture, which converts the video into a latent matrix representation. (2) The diffusion process, where seeded noise is added to the latent matrix (latent seed) which is denoised by a spatio-temporal UNet conditioned by the latent representation obtained from the audio encoder. (3) The decoding process, which used the decoder of the pretrained VAE to generate the final output. Notice that the video encoding is removed during inference.}
    \label{fig:metho}
\end{figure}

Figure~\ref{fig:metho} illustrates the steps and architecture employed in this work. The process begins by encoding the training video frames, $f_{\text{gt}}^i$, into their corresponding latent representations, $\mathbf{Z}_{\text{gt}}^0$, using the VAE encoder $E$. As shown in Eq.~\ref{eq:latents},  each frame is downsampled, reducing its spatial dimensions to $H/8$ and $W/8$ (height and width, respectively). The resulting latent representation consists of 4 channels, a design choice aimed at optimizing computational efficiency while preserving essential information.

\begin{equation}
    Z^{0}_{\text{gt}} = [E(f_{gt}^1), \dots, E(f_{gt}^F)] \in \mathbb{R}^{F \times H/8 \times W/8 \times 4}
    \label{eq:latents}
\end{equation}
\newline

During the training phase, the diffusion process incrementally introduces Gaussian noise $[\epsilon_{gt}^1, ..., \epsilon_{gt}^T]$ to the ground truth ($gt$) over $T$ timesteps, transforming $Z_{gt}^0$ into $Z_{gt}^T$. As a result, the sequence $[Z_{gt}^0, \ldots, Z_{gt}^T]$ gradually contains less information with each step forward in the diffusion process.
In exploring noise addition strategies for latent representations, we compared Denoising Diffusion Probabilistic Models (DDPMs)~\cite{ho2020denoising} and Pseudo Numerical methods for Diffusion Models (PNDMs)~\cite{liu2021pseudo}. DDPM is a linear predefined scheduler, which offers simplicity and ease of implementation but potentially sacrificing flexibility, efficiency, and output quality. Conversely, PNDMs adopt a learned noise schedule, which dynamically adjusts noise levels based on data and the current generation phase, which make it more adaptable to the distribution of the training data
The training process utilizes the loss function in Eq.~(\ref{eq:losses}), which consists of two parts: the first is the Mean Squared Error (MSE) for reconstruction accuracy, and the second is the temporal coherence loss, ensuring smooth frame transitions between latent representations. Here, the term $f_{t}^i$ indicates the $i$-th  frame of the noise-corrupted sequence at time $t$, while $f_{gt}^i$ is the $i$-th frame of the original video. The weights $\lambda_{1}$ and $\lambda_{2}$ adjust the impact of the MSE reconstruction and temporal coherent losses, respectively, ensuring the generated video maintains both visual fidelity to the original content and fluidity across frames as follows:

\begin{equation}
    \begin{split}
        \mathcal{L}(\theta) = \lambda_{1} \sum_{i=1}^{F} \| D(E(f_\text{nt}^i)) - f_\text{gt}^i \|^2 + \\ 
        \lambda_{2} \sum_{i=2}^{F} \| (E(f_\text{nt}^i) - E(f_\text{nt}^{i-1})) - (E(f_\text{gt}^i) - E(f_\text{gt}^{i-1})) \|^2
        \end{split}
    \label{eq:losses}
\end{equation}

In inference, the UNet ($\epsilon_{\theta}$) predicts the added noise at each step, generating $Z_{pr}^0 = [z_{pr}^1, \ldots, z_{pr}^F]$ from the initial random noise $Z_{pr}^T$, as shown in Eq.~\ref{eq:syn}. $\tau(y)$ is the embedding obtained from the Wav2Vec model for a given speech segment $y$, ensuring that the generated video aligns with the intended semantic content. Subsequently, the VAE decoder~$D$ transforms these latent representations $Z_{pr}^0$ directly into the final video frames.

\begin{equation}
    z_{pr}^0 = \epsilon_{\theta}(z_{pr}^T, T, \tau(y))
    \label{eq:syn}
\end{equation}
\newline
At the inference stage, the UNet predicts the added noise at each step, generating $Z_{pr}^0 = [z_{pr}^1, \ldots, z_{pr}^F]$ from the initial random noise $Z_{pr}^T$. The embedding obtained from the audio encoder ensures that the generated video aligns with the intended semantic content. Subsequently, the VAE decoder transforms these latent representations $Z_{pr}^0$ directly into the final video frames. This method enables the creation of videos with high temporal coherence and fidelity to the audio input, mirroring the dynamics seen in natural speech-driven video content.


\section{Experiments \& Results}

For evaluation purposes, we used two distinct datasets that provide RT- and cine-MRI visualizations of the vocal tract. The first database is the publicly available \textit{USC 75-Speaker Speech MRI Database}~\cite{lim2021multispeaker}, which is a collection of 2D sagittal-view RT-MRI videos, paired with synchronized audio recordings. For the second in-house dataset, we acquired paired 2D cine-MRI sequences and audio recordings from 30 subjects (10 with tongue cancer and 20 healthy controls).

For comparison, we also used a VQGAN and adapted a DiffAE with conditioning as our Spatial Diffusion (SDiff) model. For the VQGAN, we employed a methodology inspired by the architecture proposed in~\cite{esser2021taming}. The model adapted here aims to translate speech audio signals into MRI images by encoding audio inputs with Wav2Vec and transforming these into 2D formats for the generator, which reconstructs MRI images. In addition, the DiffAE architecture is based on the one proposed in~\cite{preechakul2022diffusion}, which integrates a semantic encoder based on a ResNet which also incorporates a denoising diffusion model to achieve decodable and semantically meaningful image representations. 

Our proposed framework includes a dual latent code for capturing both semantic meaning and fine details, incorporating Wav2Vec's speech representations, contrastive training, and noise prediction enhancements to refine image reconstruction fidelity. We adopted a stable diffusion approach for our STDiff model, inspired by the text-to-video framework described in~\cite{rombach2022high}, to transform speech into coherent videos. This approach integrates spatial and temporal dynamics, ensuring seamless frame transitions that correspond to the speech audio cues. We used a conditional 3D-UNet that processes these dynamics across video frames. For conditioning, we employ audio embeddings (Wav2Vec) to refine noise into structured video frames via a diffusion process. 

We performed both quantitative and qualitative analyses. The proposed approach was evaluated using the Fréchet Video Distance (FVD), Kernel Inception Distance (KID), Structural Similarity Index Measure (SSIM), and Peak Signal-to-Noise Ratio (PSNR). For FVD and KID metrics, lower scores indicate better performance, whereas for SSIM and PSNR, higher scores indicate improved quality. In the qualitative/human evaluation, we conducted a survey with 30 participants from different backgrounds and nationalities. 

\subsection{Quantitative Analysis}
In our study, we performed three main experiments to evaluate the performance of our model in capturing/reproducing the dynamics of vocal tract movements during speech: (1) Using speech from an unknown participant, (2) removing certain words from the training data and then evaluated the model's performance when predicting these omissions, and (3) in a second dataset, which comprises both Tongue Cancer (TC) patients and healthy controls.
Initially, we pretrained our models on a subset of 69 participants from the \textit{USC 75-Speaker} dataset. Following pretraining, we fine-tuned the models on recordings from six different individuals (3 female, 3 male), excluding tasks that did not involve speech. The objective was to determine whether our approach could accurately model the dynamics of the vocal tract movements of a speaker.

 For the unknown speaker task, we trained the model on entire recordings and tested it on audio from other participants who performed identical tasks. This process evaluated the model's ability to generalize speech dynamics from learned phonemes and speech information, and its capacity to translate this knowledge into video using a speaker-specific model.
The quantitative results of this experiment are shown in Tab.~\ref{tab:span}, where as first step we compared the performance of the different tested models.

\begin{table}[!htbp]
\centering
\caption{Evaluation of Different Models on the \textit{USC 75-Speaker Database} for Generalizing Vocal Tract Dynamics with Unknown Speaker and Excluded Words During Training}
\label{tab:span}
\centering
\setlength{\tabcolsep}{12pt}
\resizebox{\linewidth}{!}{
\begin{tabular}{ccccc}
\hline
\textbf{Model} & \textbf{KID ($\downarrow$)}      & \textbf{FVD ($\downarrow$)}                                      & \textbf{SSIM ($\uparrow$)}       & \textbf{PSNR ($\uparrow$)}        \\ \hline
\multicolumn{5}{c}{\textbf{Unknown Speaker}}                                                                                                                                                \\ \hline
VQGAN       & 0.15\,$\pm$\,0.04  & \phantom{1}984.46\,$\pm$\,419.78  & 0.66\,$\pm$\,0.07 & 21.46\,$\pm$\,2.19 \\
SDiff      & 0.16\,$\pm$\,0.03 & 1062.78\,$\pm$\,372.04 & 0.79\,$\pm$\,0.01 & 20.51\,$\pm$\,0.40 \\
STDiff$_{D}$& 0.12\,$\pm$\,0.08  & \phantom{1}612.62\,$\pm$\,126.05  & 0.74\,$\pm$\,0.02 & 23.00\,$\pm$\,1.50 \\
STDiff$_{P}$ & 0.08\,$\pm$\,0.01  & \phantom{1}538.72\,$\pm$\,132.42  & 0.81\,$\pm$\,1.11 & 24.06\,$\pm$\,3.55 \\ 
 \hline
\multicolumn{5}{c}{\textbf{Unknown Words}}  \\ \hline
VQGAN    & 0.20\,$\pm$\,0.02  & 1066.60\,$\pm$\,297.82  & 0.80\,$\pm$\,0.02   & 23.26\,$\pm$\,0.33  \\
SDiff    & 0.13\,$\pm$\,0.02  & \phantom{1}921.03\,$\pm$\,382.31 & 0.77\,$\pm$\,0.03   & 22.16\,$\pm$\,1.12  \\ 
STDiff$_{D}$   & 0.08\,$\pm$\,0.02  & \phantom{1}800.42\,$\pm$\,172.66  & 0.75\,$\pm$\,0.02   & 23.51\,$\pm$\,0.61  \\
STDiff$_{P}$   & 0.11\,$\pm$\,0.03  & \phantom{1}702.94\,$\pm$\,219.85 & 0.75\,$\pm$\,0.02   & 23.43\,$\pm$\,0.71  \\  \hline
\multicolumn{5}{l}{\footnotesize{STDiff$_{P}$: STDiff using PNDM. STDiff$_{D}$: STDiff using DDPM.}}                                                                          
\end{tabular}}
\end{table}

\begin{figure}[!hb]
    \centering
    \includegraphics[width = .8\textwidth]{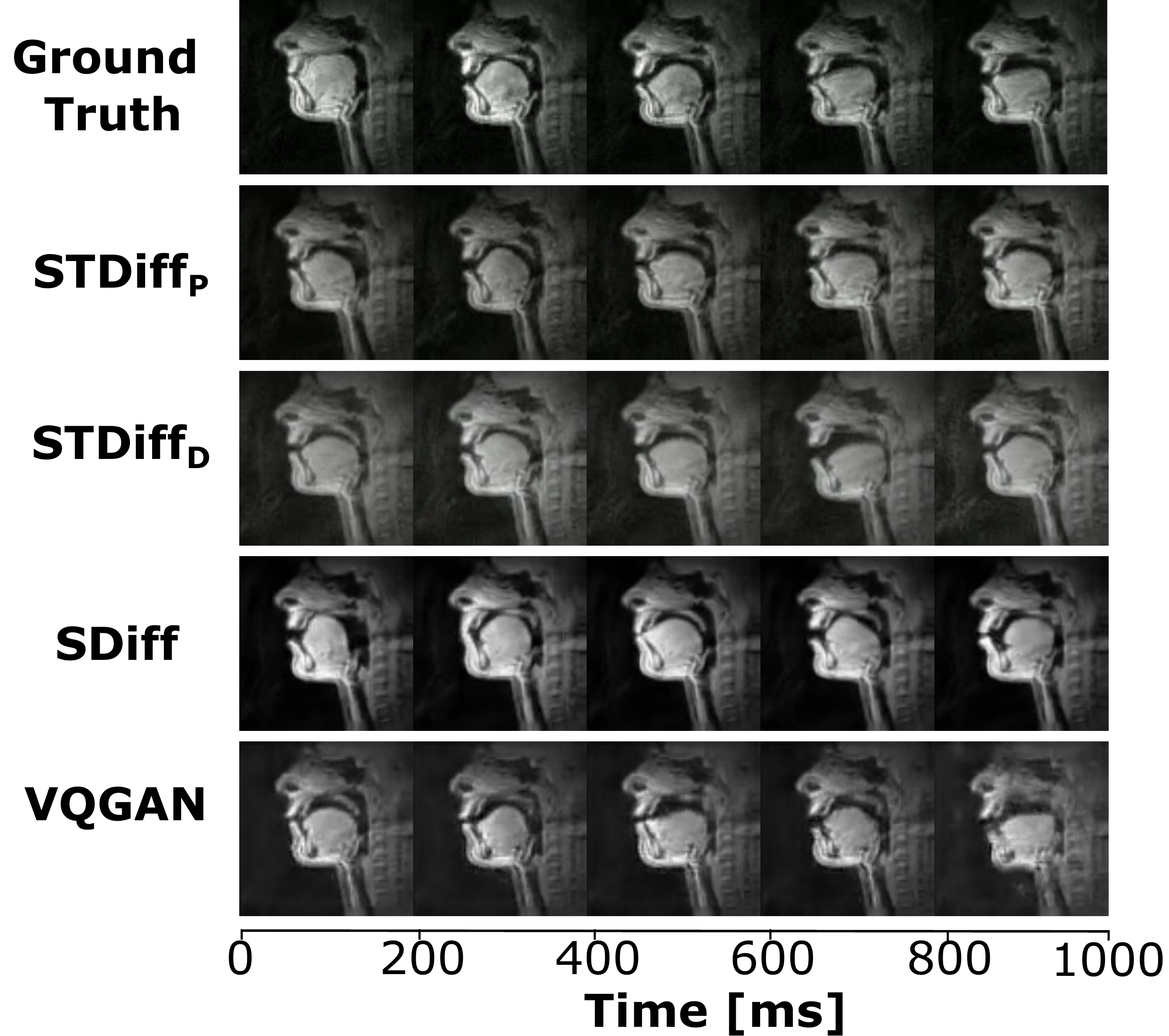}
    \caption{Comparison of the models used in this work for a female participant when saying ``When the''. The videos were resample to 5\,\textit{fps} only for demonstration purposes.}
    \label{fig:ex1}
\end{figure}

STDiff models obtained the highest and more consistent performance when simulating the dynamics of the vocal track in unknown speaker. An example of the generated videos is displayed in Fig.~\ref{fig:ex1}.

The second experiment focused on the model's performance when certain words, such as ``sun'', ``light'', ``sunlight'', ``rain'', ``bow'', and ``rainbow'', were omitted from the training data. 
These words were selected based on their potential for varied pronunciation or production. The testing phase involved recordings from a speaker different from those used in the first experiment, to evaluate the model's capacity to generate accurate videos based on the speaker's ground truth.

The overall performance of the models can be found in Tab.~\ref{tab:span}, where they follow a trend similar to that of the STDiff-PNDM based model in obtaining more accurate results. However, we can observe that, according to the SSIM, the frames are more dissimilar, which was expected since that exact combination of sounds was hardly or never seen during training. A possible application in this area could offer a basis for novel systems for physicians to understand the dynamics of a patient's speech without the need for the patient's physical presence, while simulating it vocal tract behavior to certain sounds.

The final experiment aims to explore the model's use in clinical settings for tongue cancer patients, using data from participants with reduced tongue mobility after glossectomy. Our goal was to evaluate whether the model could still effectively capture speech dynamics despite the physical limitations imposed by the condition. 
 We can observe that this test outperformed the one in the \textit{USC-75 speaker} data (see Tab.~\ref{tab:mary}).

\begin{table}[!htbp]
\centering
\caption{Evaluation of S2MRI Models on Tongue Cancer Patients and Healthy Controls for Generalizing Vocal Tract Dynamics with Unknown Speaker }
\label{tab:mary}
\setlength{\tabcolsep}{12pt}
\resizebox{\linewidth}{!}{
\begin{tabular}{ccccc}
\hline
\textbf{Model} & \textbf{KID ($\downarrow$)}      & \textbf{FVD ($\downarrow$)}                                      & \textbf{SSIM ($\uparrow$)}       & \textbf{PSNR ($\uparrow$)}        \\ \hline
\multicolumn{5}{c}{\textbf{TC Patients}}                                                                                                                                                 \\ \hline
VQGAN   & 0.46\,$\pm$\,0.02  & 1766.61\,$\pm$\,1456.63  & 0.76\,$\pm$\,0.02   & 16.53\,$\pm$\,0.04  \\
SDiff    & 0.53\,$\pm$\,0.02  & 1616.62\,$\pm$\,30.08\phantom{11} & 0.74\,$\pm$\,0.04   & 17.15\,$\pm$\,0.18 \\ 
STDiff$_D$  & 0.20\,$\pm$\,0.09  & 306.10\,$\pm$\,120.14  & 0.78\,$\pm$\,0.04   & 23.81\,$\pm$\,0.52  \\
STDiff$_P$     & 0.19\,$\pm$\,0.09  & 303.55\,$\pm$\,\phantom{1}98.22  & 0.83\,$\pm$\,0.05   & 25.21\,$\pm$\,0.73  \\ 
 \hline
\multicolumn{5}{c}{\textbf{Healthy Controls}}     \\ \hline
VQGAN    & 0.34\,$\pm$\,0.02  & 1118.58\,$\pm$\,300.34\phantom{1} & 0.69\,$\pm$\,0.02   & 15.34\,$\pm$\,0.04   \\
SDiff    & 0.12\,$\pm$\,0.01  &  926.67\,$\pm$\,320.44  & 0.62\,$\pm$\,0.02   & 18.94\,$\pm$\,0.25  \\ 
STDiff$_D$     & 0.22\,$\pm$\,0.07  & 280.07\,$\pm$\,153.35 & 0.76\,$\pm$\,0.08   & 23.16\,$\pm$\,1.30   \\
STDiff$_P$     & 0.22\,$\pm$\,0.06  & 269.63\,$\pm$\,134.64  & 0.82\,$\pm$\,0.06   & 23.51\,$\pm$\,1.15  \\   \hline
\multicolumn{5}{l}{\footnotesize{STDiff$_{P}$: STDiff using PNDM. STDiff$_{D}$: STDiff using DDPM.}}                                                                          
\end{tabular}}
\end{table}

\begin{figure}[!ht]
    \centering
    \includegraphics[width = 0.8\linewidth]{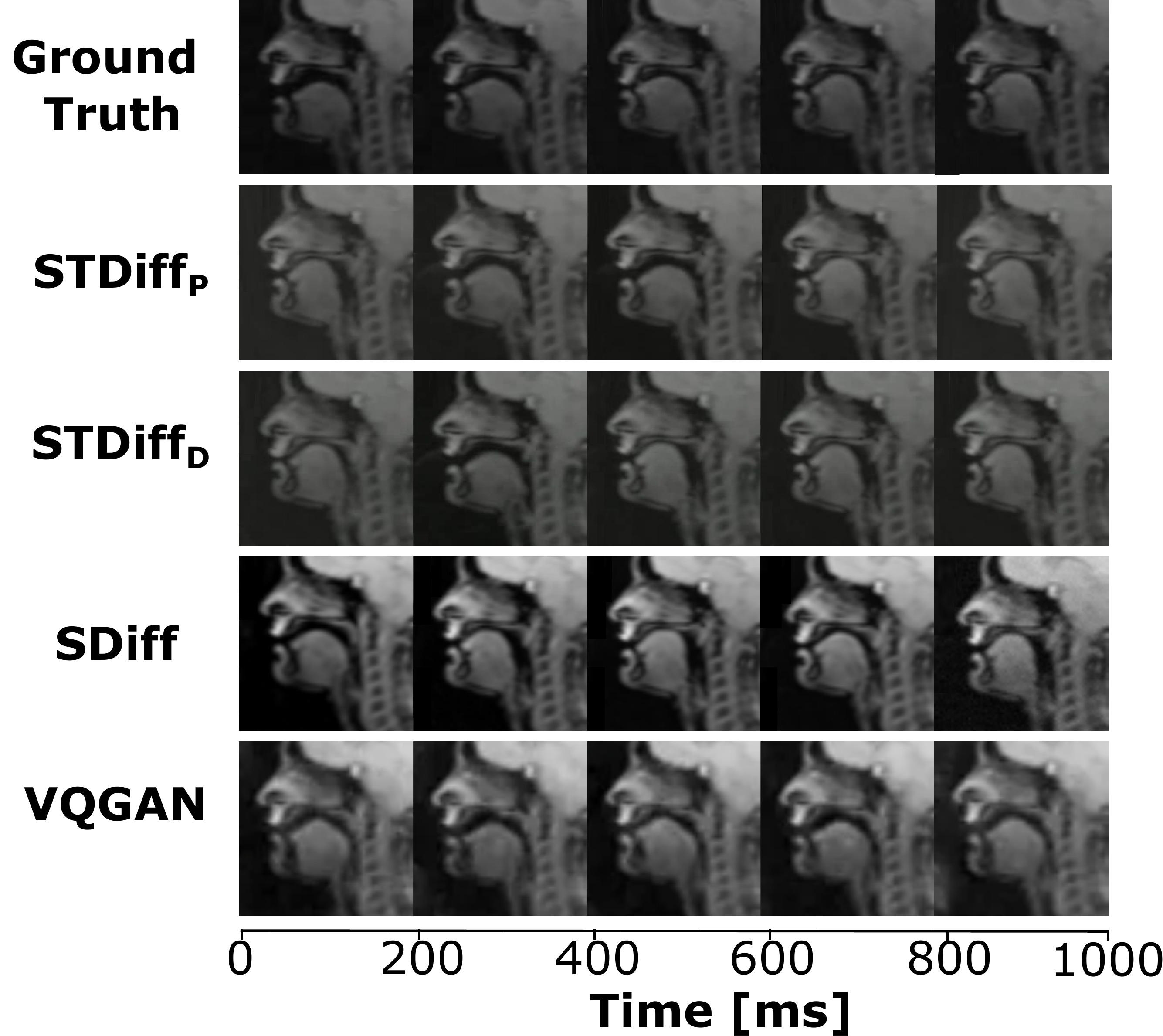}
    \caption{Comparison of the models employed in this work for a patient pronouncing the word ``a souk''. The videos were resample to 5\,\textit{fps} only for demonstration purposes.}
    \label{fig:ex2}
\end{figure}

However, one reason may be that it is fine-tuned on shorter speech segments, which only captures the dynamic from only two words. Notice that the system was able to get closer to the video of the healthy controls than the patients. This could be due to the pretraining only using healthy controls.
The qualitative results for a patient are shown in Fig.~\ref{fig:ex2}.

\section{Human evaluation}
We conducted a survey in which 30 participants provided basic demographic information and ranked nine videos according to a multi-answer questionnaire designed for the purpose of evaluating generative AI, as described in~\cite{otani2023toward}. The participants ranged from phoneticians and linguists to computer scientists and engineers, from different countries and backgrounds.
Tab.~\ref{tab:quali} presents a comparative analysis of viewer perceptions regarding the authenticity and audio-visual alignment of videos, categorized across three different sources. The first section is related to the audience's perception of the videos' sources, while the second focuses on evaluating how well the video content matches the audio. The VQGAN was excluded due to its inability to produce smooth transitions, and none of the samples generated from longer words accurately corresponded to the audio content. Both schedulers from the STDiff model exhibited similar results, leading us to include them in a single category to avoid extending the survey.

\begin{table}[!htpb]
\centering
\caption{Comparison of viewer perceptions regarding video authenticity and audio-visual alignment among different generative models and real videos. Values represent the average percentage of participants selecting each option}
\label{tab:quali}
\setlength{\tabcolsep}{4pt} 
\resizebox{\linewidth}{!}{
\begin{tabular}{lccc}
\hline
\textbf{Option} & \textbf{GT} & \textbf{S--Diff} & \textbf{ST--Diff} \\ 
\hline
\multicolumn{4}{c}{\textbf{Does the video look like an AI-generated video or a real video?}} \\ 
\hline
AI-generated video & 17.23 & 13.75 & 25.27 \\
\begin{tabular}[c]{@{}l@{}}Probably an AI generated video, but photorealistic\end{tabular} & 29.90 & 39.65 & 27.60 \\
Neutral & 18.37 & 22.40 & 21.83 \\
\begin{tabular}[c]{@{}l@{}}Probably a real video, but with irregular textures/shapes\end{tabular} & 17.23 & 10.35  & 6.87 \\
Real video & 20.67 & 13.80 & 18.40 \\ 
\hline
\multicolumn{4}{c}{\textbf{How well does the video match the audio?}} \\
\hline
Does not match at all & 3.43 & 18.95 & 13.77 \\
Has significant discrepancies & 13.77 & 22.40 & 17.23 \\
Has several minor discrepancies & 12.60 & 17.25 & 18.37 \\
Has few minor discrepancies & 39.07 & 18.95 & 28.73\\
Matches exactly & 29.87 & 22.40 & 21.83\\
\hline
\multicolumn{4}{l}{GT: Ground Truth. S--Diff: Spatial Diffusion. ST--Diff: Spatio-Temporal}\\
\multicolumn{4}{l}{Diffusion. Values are expressed in percentages [\%]}
\end{tabular}}
\end{table}

The synchronization between audio inputs and visual MRI outputs was generally impressive, with most participants noting only minor discrepancies. This underscores the models' capability to effectively capture and visualize audio descriptions. Despite some notable exceptions where significant discrepancies emerged, these instances underlined the opportunity to refine complex speech-to-visualization translations.

Participants highlighted technical areas needing enhancement, such as the ST--Diff model's challenges in rendering fine details and the S--Diff model's inconsistency across sequences. Nevertheless, the S--Diff model was often perceived as more photorealistic, suggesting its superior handling of textures and details which enhances its realism, while the ST--Diff was perceived better to handle transitions.

Notable feedback included an expert linguist's observation on the DiffAE (S--Diff) model, where there was a misalignment in articulatory features critical for accurate phoneme visualization--``The last phoneme sounds like a `t', but the tongue doesn't touch the teeth'' and ``The tip of the tongue sometimes looks weird for the sound 'th', which does not match perfectly''.
Despite these challenges, the positive response to most generated sequences highlights their substantial potential for applications in educational tools, clinical training, and preliminary diagnostics, where precision is less critical.

The survey results revealed mixed perceptions of authenticity, with some AI-generated videos being mistaken for real footage. This indicates the models' advanced capabilities in mimicking reality, though some artificial elements were still detectable. These results not only demonstrate the models' application potential in fields requiring high fidelity but also highlight the need for ongoing improvements to enhance realism consistently.

Overall, these insights remind us of the subjective nature of visual perception and the varied sensitivity among viewers in distinguishing real from synthetic imagery, which could guide future enhancements in model training.


\section{Discussion \& Conclusions}

In our work, we proposed a framework for Speech-to-Video translation, which uses generative models based on stable diffusion to transform speech signals into dynamic cine-MRI visualizations of the vocal tract.

Our framework demonstrated adaptability to unknown speech and to generalize from omitted words. By employing textless generation techniques, we offer a novel solution aimed at easing the challenges encountered by both patients and therapists. The quantitative results indicated that the STDiff model, particularly when using the PNDM scheduler, outperformed other models in terms of lower FVD and KID, highlighting its superior capability in producing coherent video sequences that closely mimic the dynamics of actual speech movements.
This model's ability to integrate spatial and temporal dynamics ensures that the video frames not only match well with the audio cues but also exhibit seamless transitions, making it particularly useful for applications requiring detailed and dynamic visual representations such as in clinical settings for speech therapy or surgical planning.

In comparison, SDiff and VQGAN models showed varied results. The SDiff model generally performed well, offering realistic textural details that enhance the photorealism of the generated images. However, it occasionally struggled with maintaining consistency across sequences, which is crucial for clinical applications where precise and reliable visual outputs are necessary. On the other hand, the VQGAN, although capable of generating high-quality images, faced challenges in producing smooth transitions between frames, which is a critical aspect when representing continuous speech movements.

Further, the human evaluation highlighted the model's performance, where the S--Diff model was often perceived as more photorealistic compared to the ST--Diff model.
However, in terms of handling transitions, the ST--Diff model, particularly when employing the PNDM scheduler, outperformed other models. It demonstrated superior capability in producing coherent video sequences that closely mimic the dynamics of actual speech movements
This distinction likely arises due to the complexity of the ST--Diff model complexity that aims to simultaneously achieve high fidelity in both detail and motion.

Despite encountering limitations such as modeling pauses, misalignment in predictions, and the necessity for extensive data for accurate simulation and reproduction of unknown words, the study showed promising results. 

Future work will aim to refine these models by integrating phonetic constraints to improve the accuracy of phoneme visualization and applying foundational models to reduce the heavy data demands for training. Moreover, as a potential future application, therapists could use this type of system to automatically simulate personalized vocal tract dynamics for outpatient analysis.
%


\section*{CRediT authorship contribution statement}

\textbf{P. A. Pérez-Toro:} Conceptualization, Methodology, Software, Data curation, Formal analysis, Writing—original draft, Writing—review and editing, Visualization. \textbf{T. Arias-Vergara:} Conceptualization, Writing—review and editing, Visualization. \textbf{F. Xing: } Conceptualization, Writing—review and editing. \textbf{X. Liu:} Conceptualization, Writing—review and editing. \textbf{M. Stone:} Writing—review and editing. \textbf{J. Zhuo:} Writing—review and editing. \textbf{J. R. Orozco-Arroyave:} Writing—review and editing.
\textbf{E. Nöth:} Conceptualization, Writing—review and editing. \textbf{J. Hutter:} Writing—review and editing. \textbf{Jerry L. Prince:} Writing—review and editing. \textbf{A. Maier:} Conceptualization, Writing—review and editing. \textbf{J. Woo:} Conceptualization, Writing—review and editing.

\section*{Declarations}

\subsection*{Competing interests}
The authors declare no competing interests.

\section*{Data Availability}
 For additional details about the corpora used in this study and to request access, please refer to  \url{https://sail.usc.edu/span/75speakers/} in the case of the USC 75-Speaker Speech Real Time MRI Database. Due to the confidential nature of the Tongue Cancer cine-MRI data, we are unable to make our data publicly available. Access to the data can be made through reasonable requests and will be subject to local ethics clearances. Please email the senior author at \url{jwoo@mgh.harvard.edu}.

\section*{Acknowledgements}
This work was partially funded by the EVUK programme (``Next-generation Al for Integrated Diagnostics'') of the Free State of Bavaria, by CODI at UdeA grant \#PI2023-58010, and the NIH grants \#DC018511, \#R01DC014717 and \#R01CA133015. The authors gratefully acknowledge the scientific support and HPC resources provided by the Erlangen National High Performance Computing Center (NHR@FAU) of the Friedrich-Alexander-Universität Erlangen-Nürnberg (FAU). The hardware is funded by the German Research Foundation (DFG).

%


\bibliographystyle{elsarticle-num-names} 

\end{document}